\documentclass[preprint,12pt]{elsarticle}

\usepackage{amsmath,amssymb,amsfonts}
\usepackage{booktabs}
\usepackage{hyperref}
\usepackage{graphicx}
\graphicspath{{doc/image/}}
\begin{document}

\begin{frontmatter}



\title{Investigating the Influence of Prompt and Response Languages on LLM Content Generation}


\author[label2]{Thi Thanh Nhan Nguyen$^*$}            
\author[ntnu]{Mai Khoi Tieu$^*$\footnote{$^*$denotes equal contribution}}
\author[label1]{Michael A. Riegler}
\author[label1]{Pål Halvorsen}
\author[hutech]{Thu Nguyen}

\affiliation[ntnu]{organization={Norwegian University of Science and Technology (NTNU)},
            city={Trondheim},
            % postcode={7034},
            state={Trøndelag},
            country={Norway}}
\affiliation[label1]{organization={SimulaMet},
            city={Oslo},
            country={Norway}}
\affiliation[label2]{organization={Université de Technologie de Compiègne},
            % addressline={Rue du Docteur Schweitzer CS 60319},
            city={Compiègne},
            % postcode={60203},
            state={Hauts-de-France},
            country={France}}
% ,, Viet Nam
% nt.thu93@hutech.edu.vn
            
\affiliation[hutech]{organization={Faculty of Information Technology, HUTECH University},%Department and Organization
            city={ Ho Chi Minh City},
            country={Vietnam}}

\begin{abstract}
This study investigates how the choice of prompt and response language shapes the generative behavior of Large Language Models (LLMs), extending beyond surface-level linguistic variation to examine structural and semantic dimensions of output. Using five models---DeepSeek V3, GPT-4o, Phi-4-multimodal, Claude 3.5 Haiku, and Gemini 2.5 Pro---we evaluated responses to 68 non-translation questions spanning ethics, culture, health, and social domains under four conditions defined by prompt$\times$response language: English$\rightarrow$English, English$\rightarrow$Norwegian, Norwegian$\rightarrow$Norwegian, and Norwegian$\rightarrow$English. After excluding items refused in any condition, our balanced dataset comprises 1{,}348 responses (337 per condition). Length differences were quantified with Cohen's $d$; semantic fidelity with LaBSE cosine similarity (chunk-averaged to avoid truncation); and cross-lingual keyword overlap with both raw Jaccard and a LaBSE-based soft Jaccard that maps Norwegian and English keywords into a shared embedding space. We identify a robust prompt-language effect on response length that is visible in the confound-free within-response-language contrasts: holding the response language fixed at English, a Norwegian prompt shortens responses by 37\% on average ($ee$ 427.8 vs.\ $ne$ 268.2 words, $d\approx 0.94$); holding the response language fixed at Norwegian, an English prompt shortens responses by 41\%.
%($nn$ 344.6 vs.\ $en$ 203.8, $d\approx 0.95$). 
The largest single-cell, cross-lingual $ee$--$en$, is 52\% shorter in words ($d\approx 1.32$), but only 25\% shorter in tokens---so a portion of that headline figure reflects Norwegian's higher subword-tokenizer fertility rather than pure model compression. Cosine similarity remains high across all conditions ($\approx 0.83$, unchanged under chunk-averaged embedding), while soft Jaccard shows substantial concept overlap ($\approx 0.53$) that is invisible to raw-string Jaccard ($\approx 0.02$)---direct evidence of conceptual paraphrasing rather than literal translation. Information density is a near-perfect monotonic inverse of length and is reported as a companion, not an independent effect. Per-model Cohen's $d$ ranges widely (Phi-4 $\approx 0.78$ to GPT-4o $\approx 4.42$), so the pooled effect is heterogeneous across models. Prompt language is not a neutral parameter: it shapes output length and lexical realization, which matters for multilingual prompt engineering.
\end{abstract}


\begin{highlights}
    \item \textbf{Prompt Language Modulates Output Length:} The language of the prompt exerts a significant structural influence on the length of the generated response, independently of the specified output language.
    \item \textbf{Cross-Lingual Compression Effects:} Prompting in English for a Norwegian response yields the most pronounced text compression (a 52\% reduction in word count); however, this phenomenon is partially attributable to tokenizer fertility disparities rather than purely generative compression.
    \item \textbf{Semantic Fidelity via Conceptual Paraphrasing:} Despite substantial reductions in output length, semantic integrity remains highly stable across conditions. Models utilize conceptual paraphrasing rather than literal translation, maintaining high cross-lingual embedding similarity despite low surface-level lexical overlap.
    \item \textbf{Inverse Relationship with Information Density:} The structural compression observed in cross-lingual tasks corresponds to a near-perfect monotonic increase in information density, communicating equivalent semantic payloads with greater lexical efficiency.
    \item \textbf{Architectural Heterogeneity:} The magnitude of this translation compression effect varies significantly across different LLM architectures, with highly verbose models exhibiting the most pronounced deviations.
% \item Prompt language shortens responses even when the response language is held fixed: within English-only responses, a Norwegian prompt cuts length by 37\% ($ee$ vs.\ $ne$, $d\approx 0.94$); within Norwegian-only responses, an English prompt cuts length by 41\% ($nn$ vs.\ $en$, $d\approx 0.95$).
% \item Cross-lingual $ee$--$en$ shows the largest single-cell gap (52\% in words), but only 25\% in tokens---so a portion of the ``translation compression'' effect is Norwegian tokenizer fertility, not model behavior.
% \item Semantic content stays stable across conditions ($\approx 0.83$ cosine similarity, unchanged under chunk-averaged LaBSE); cross-lingual keyword overlap is $\approx 0.02$ by raw Jaccard but $\approx 0.53$ by LaBSE-embedded soft Jaccard---direct evidence of conceptual paraphrasing that raw-string comparison cannot detect.
% \item Per-model Cohen's $d$ for the $ee$--$en$ gap ranges from 0.78 (Phi-4) to 4.42 (GPT-4o); the pooled effect is heterogeneous across models.
% \item Information density is a near-perfect monotonic inverse of length (Spearman $\rho\approx -1.0$), reported as a companion metric rather than independent evidence.
\end{highlights}


\end{frontmatter}

\section{Introduction}

The deployment of Large Language Models (LLMs) in multilingual environments raises critical questions regarding how language selection influences the nature of generated content. While LLMs are capable of processing queries in various languages, it remains unclear whether the choice of language merely affects syntax or if it fundamentally alters the underlying information retrieval and presentation strategies. This study analyzes these behaviors by examining how models respond when the language of the prompt differs from the required language of the response.

To investigate this, we structured an experiment using five models---DeepSeek V3, GPT-4o, Phi-4-multimodal, Claude 3.5 Haiku, and Gemini 2.5 Pro. The experimental design involved asking identical questions across four topics (ethics, culture, health, and social) using four linguistic permutations: prompting in English or Norwegian, and requesting responses in either language.

We then analyzed the responses along four axes. First, we measured the magnitude of the ``translation compression'' effect with Cohen's d, distinguishing systematic shifts in response length from random variation. Second, to check that this brevity did not come at the cost of meaning, we assessed content fidelity with semantic similarity, encoding responses with LaBSE embeddings and computing their cosine similarity. Third, we computed Information Density to test whether shorter outputs were more information-dense or merely truncated. Finally, we measured Lexical Divergence with the Jaccard index to distinguish literal translation from conceptual paraphrasing.
By applying these metrics, this paper provides quantitative evidence on how language constraints act as a mechanism for ``translation compression'' and cultural framing within generative AI.

In short, the contributions of this work are as follows. First, we provide a systematic evaluation of how input and output language constraints influence the generation behavior of modern LLMs, moving beyond performance benchmarks to structural and semantic shifts. Second, we introduce an analytical framework combining Cohen's $d$, LaBSE-based semantic similarity, information density, and Jaccard lexical overlap to quantify the trade-off between rhetorical elaboration and informational efficiency. Third, using a $2\times2$ design, we show that ``translation compression'' is a prompt$\times$response interaction rather than a uniform cross-lingual effect: the most compressed condition (English prompt $\rightarrow$ Norwegian response) is about 52\% shorter and roughly $2\times$ denser than English-to-English generation. Finally, we document a dissociation between semantic content---which is quantified and largely preserved ($\approx 0.83$ similarity)---and cultural presentation, which we characterize qualitatively as shifting with language choice.

The remainder of this paper is organized as follows. Section \ref{sec-related} reviews related work. Sections \ref{sec-method} and \ref{sec-eval-strategy} detail the experimental design, defining the four prompt$\times$response language conditions and the evaluation metrics: Cohen's $d$, a mixed-effects length model, LaBSE-based cosine similarity (with chunked robustness), information density, and both raw and LaBSE soft Jaccard for keyword overlap. Section \ref{sec-result} reports the quantitative results, including per-model heterogeneity, char- and token-based robustness checks for the length effect, and the semantic--lexical dissociation. Section \ref{sec-discuss} discusses candidate mechanisms (tokenizer fertility, RLHF verbosity bias, pretraining-corpus asymmetry, safety hedging) and limitations. Section \ref{sec-conclu} concludes with recommendations for multilingual prompt engineering.

\section{Related Works}\label{sec-related}

The study of multilingual Large Language Models (LLMs) has evolved from simple performance benchmarking to complex investigations into how linguistic constraints influence cognitive reasoning and cultural alignment. This section reviews three key areas of related literature to our study: the dynamics of cross-lingual prompting, relevant cultural aspects and the evolution of semantic evaluation metrics.

\subsection{Cross-Lingual Prompting and Information Compression}
Prior research has extensively documented the ``English-centric'' bias of modern LLMs, where models demonstrate superior reasoning capabilities in English due to the predominance of English data in pre-training corpora~\cite{brown2020fewshot,scao2022bloom}. Consequently, cross-lingual prompting (prompting in a high-resource language like English for output in a low-resource language) has been proposed as a strategy to unlock better reasoning capabilities in multilingual tasks~\cite{qin2023crosslingual,shi2023multilingual}. However, the structural impact of this language switching on \emph{response length} has received less attention.

While recent studies on ``prompt bloat'' suggest that excessive context can degrade reasoning~\cite{liu2024lostmiddle,he2024promptformat}, our observation of a ``Translation Compression'' effect aligns more closely with findings on semantic compression. Gilbert et al.~\cite{gilbert2023semantic} demonstrated that LLMs can effectively compress text while preserving semantic essence, suggesting an inherent capability to distill information when constrained. However, they studied compression as an explicit task, whereas we observe it arising on its own during cross-lingual generation. Having to answer in another language appears to act as a constraint: the model delivers the core content more concisely and drops much of the ``conversational filler'' typical of English-to-English responses.

\subsection{Linguistic Determinism and Cultural Framing}
The hypothesis that the language of the prompt influences the cultural values and generation style of the output---a computational parallel to the Sapir-Whorf hypothesis---has gained empirical support~\cite{lu2025cultural,hershcovich2022crosscultural}. Li et al.~\cite{li2024culturellm} introduced ``CultureLLM'' to address the Western bias inherent in English-prompted generations, noting that standard alignment techniques often fail to capture local cultural nuances. Similarly, varying the prompt language has been shown to shift model outputs between ``independent'' (Western) and ``interdependent'' (Eastern) social orientations, effectively toggling the model's active cultural framework~\cite{lu2025cultural}.

Our results are consistent with this line of work in the length and lexical-realization channels: outputs shift systematically with the language of instruction. We do not, however, quantify cultural framing directly in this study---we return to that limitation in Section~\ref{sec-discuss}.

\subsection{Semantic Evaluation in Multilingual Contexts}
Evaluating the quality of cross-lingual generation requires metrics that transcend simple lexical overlap. Traditional n-gram metrics like BLEU~\cite{papineni2002bleu} or ROUGE~\cite{lin2004rouge} have proven insufficient for capturing semantic fidelity in open-ended generation, particularly when the output length varies significantly~\cite{sai2022metrics}. The shift toward embedding-based metrics, such as BERTScore~\cite{zhang2020bertscore} and LaBSE (Language-agnostic BERT Sentence Embeddings)~\cite{feng2022labse}, allows for the quantification of meaning preservation across languages regardless of syntactic structure.

Our methodology adopts these advanced metrics but integrates them with \emph{Information Density} analysis. While high semantic similarity (via LaBSE) confirms content preservation, it does not measure communicative efficiency. By combining semantic stability with density metrics, we provide a more granular view of how LLMs trade off rhetorical elaboration for informational compactness during cross-lingual tasks.
\section{Methodology}\label{sec-method}

This section describes the question set, the four language conditions, the models queried, and how we handled refusals and translation-task items.

\paragraph{Question set} Questions span four topics: ethics, culture, health, and social. The 20 culture questions were generated by Tulu~AI; the 20 ethics and 20 social questions were generated by Microsoft Copilot; the 10 health questions were written manually by the authors to include a realistic mix of first-person clinical queries. Provenance is inconsistent across topics; we treat this as a limitation (Section~\ref{sec-discuss}) rather than a controlled variable, and have not tested whether it systematically affects the reported metrics. The health topic is smaller (10 vs.\ 20) because two of the ten items (Q9 and Q10) are translation tasks (``Please translate this medical report excerpt into Norwegian\ldots'') that make the response language over-determined and were refused by every model in both Norwegian-prompt conditions; we exclude them from the length and density analyses (see below) and report them separately in \ref{appendix-translation}. This leaves 68 non-translation questions per model. English question lists are in \ref{appendix-questions}; Norwegian translations were produced in Bokmål by a native Norwegian speaker on the author team and spot-checked against the English source for semantic equivalence. The verbatim prompt templates used for each of the four conditions are given in \ref{appendix-prompts}.

\paragraph{Language conditions} Each question is issued under four conditions defined by prompt language $\times$ required response language:
\begin{itemize}
    \item $ee$: English prompt $\to$ English response
    \item $en$: English prompt $\to$ Norwegian response
    \item $nn$: Norwegian prompt $\to$ Norwegian response
    \item $ne$: Norwegian prompt $\to$ English response
\end{itemize}
The short labels ($ee, en, nn, ne$) are internal shorthand; we italicize them throughout to distinguish them from ISO~639-1 language codes.

\paragraph{Models and decoding} We queried five models: DeepSeek~V3, GPT-4o ({2024-08-06} Azure snapshot), Phi-4-multimodal-instruct, Claude~3.5~Haiku  ({claude-3-5-haiku-latest}), and Gemini~2.5~Pro ({gemini-2.5-pro-preview-03-25}). Each condition is queried once per question (no sampling). Decoding settings were held at the API defaults on Azure for GPT-4o, DeepSeek, and Phi-4 (\texttt{max\_tokens}$=1000$); Claude used \texttt{max\_tokens}$=2000$, \texttt{temperature}$=1$; Gemini used the SDK defaults. Because generation was capped, we audited the outputs for truncation and report the results in \ref{appendix-truncation}; a non-trivial fraction of responses---concentrated in Norwegian-output conditions rather than long English ones---terminate mid-sentence, which we discuss as a limitation in Section~\ref{sec-discuss}.

\paragraph{Refusals and balanced design} Three items were declined by content filters on some models: ethics~Q13 (\emph{``Is it ethical to genetically modify humans?''}) was refused by GPT-4o, DeepSeek, and Phi-4 in the Norwegian-prompt conditions, and health~Q9/Q10 (the translation-task items) were refused by every model in both Norwegian-prompt conditions. For all analyses below, we drop any (model, question) pair for which \emph{any} of the four conditions is missing, giving a balanced dataset of 337 questions~$\times$~4 conditions $=$ 1{,}348 responses. Per-model refusal counts are in \ref{appendix-refusals}.
% 1a: en, 1b: ee, 1c: nn, 1d: ne
% We asked questions in four topics: ethics, culture, health, and social. 20 questions on cultures are generated by Tulu; the sets of questions for ethics and social topics are generated by Copilot. Meanwhile, the health-related questions are manually created. The English versions of the question lists are as indicated in the following subsections.

\section{Evaluation Strategies}\label{sec-eval-strategy}

\subsection{Evaluating Response Length}
We quantify length effects with three complementary measures: word count, character count (both after light preprocessing that strips Markdown emphasis markers), and token count under a single reference tokenizer (\texttt{tiktoken} \texttt{cl100k\_base}). Reporting all three matters because Norwegian is a compounding language (e.g., \emph{informasjonstetthet} $=$ ``information density''---one word vs.\ two), and English-centric subword tokenizers have higher fertility on Norwegian, so the same content can look shorter in words but longer in tokens.

Effect sizes are reported as Cohen's $d$~\cite{cohen1988power}: $d = (\bar{x}_1 - \bar{x}_2)/s_{\text{pooled}}$, with $s_{\text{pooled}}$ the weighted pooled standard deviation. We interpret $|d|\ge 0.8$ as a large effect, $|d|\ge 0.5$ as medium, and $|d|\ge 0.2$ as small~\cite{cohen1988power}. Because the same 68 questions are asked of every model under every condition, the 1{,}348 responses are not independent; a fixed-effects ANOVA would over-estimate significance. We therefore fit a mixed-effects model with random intercepts for question:
$$
\text{length}_{ijk} \sim \text{prompt}_i \times \text{response}_j + \text{model}_k + (1 \mid \text{question}).
$$
Model is treated as a fixed factor (only five levels). We report the interaction Wald $z$-statistic and per-model $d$ values so that per-model heterogeneity is visible rather than hidden in a pooled estimate.

\subsection{Semantic Similarity Analysis}
To verify that content is preserved across languages despite length differences, we encode each response with LaBSE (Language-agnostic BERT Sentence Embeddings)~\cite{feng2022labse} into a shared multilingual space and compute cosine similarity~\cite{salton1983introduction} between response vectors. Because LaBSE truncates inputs beyond its default token window, long English responses ($ee$ frequently exceeds 500 words) could otherwise be embedded from only a prefix. As a robustness check, we also compute a \emph{chunk-averaged} variant: each response is split into 250-word windows, each window is embedded independently, and the L2-normalized mean of the window embeddings is used in place of the naive embedding. We report both.

For each question with all four conditions present, we take the mean of the six pairwise cosine similarities among $\{ee, en, nn, ne\}$ as the per-question semantic similarity. We do not impose a numerical threshold; instead we report the full distribution and, where useful, note the fraction of pairs exceeding conventional levels (e.g., $\ge 0.80$).

\subsection{Information Density}
As a companion length--vocabulary summary, we compute Information Density: the ratio of the number of TF-IDF-extracted keywords~\cite{sparckjones1972idf,salton1988termweighting} to the total word count, expressed per 100 words. Keywords are capped at 30 per response and extracted with sentence-level TF-IDF (\texttt{ngram\_range}$=(1,2)$, no stop-word filter for multilingual support). The density $\rho$ is
$$
\rho = \frac{N_{\text{keywords}}}{N_{\text{words}}} \times 100.
$$
Because the keyword count saturates near its cap for all but the shortest responses, $\rho$ is a near-perfect monotonic inverse of length in our data (Spearman $\rho_{\text{sp}}\approx -1.0$). We therefore treat it as a descriptive companion to the word-count analysis rather than an independent measure of communicative efficiency; the reader should read a higher $\rho$ as ``shorter for the same information'' rather than as an independent quality signal.

\subsection{Cross-lingual Keyword Overlap}
To characterize vocabulary reuse between response variants we report two Jaccard-style scores~\cite{jaccard1912distribution}. The \emph{raw} Jaccard is the classical set-overlap between the top-$k$ TF-IDF keywords ($K_A$, $K_B$) of two responses:
$$
J_{\text{raw}}(K_A, K_B) = \frac{|K_A \cap K_B|}{|K_A \cup K_B|}.
$$
When $A$ and $B$ are in different languages, this measure is dominated by the fact that Norwegian and English use different word forms (\emph{helse} vs.\ \emph{health}, \emph{informasjonstetthet} vs.\ \emph{information density}), so a near-zero score is guaranteed by construction and does not diagnose paraphrasing. We therefore also report a \emph{soft} Jaccard that first maps each keyword through the LaBSE embedding space and counts a keyword as matched if it has at least one nearest neighbour on the other side above a cosine threshold $\tau=0.70$:
\begin{align*}
J_{\text{soft}}(K_A, K_B) &= \frac{M}{|K_A| + |K_B| - M},
\quad M = \tfrac{1}{2}\big(m_A + m_B\big),
\end{align*}
where
\begin{align*}
m_A &= |\{a \in K_A : \max_b \cos(\phi(a),\phi(b)) \ge \tau\}|, \\
m_B &= |\{b \in K_B : \max_a \cos(\phi(a),\phi(b)) \ge \tau\}|.
\end{align*}
At $\tau = 1$ this reduces to the classical Jaccard $|K_A \cap K_B| / |K_A \cup K_B|$. Here $\phi(\cdot)$ is the LaBSE encoder. A low $J_{\text{raw}}$ paired with a high $J_{\text{soft}}$ is what we would expect from conceptual paraphrasing across languages: same concepts, different surface forms. The raw--soft gap makes the paraphrasing claim testable rather than tautological.

\section{Results and Major Remarks}\label{sec-result}

% When asking questions related to ethics:
% - content management policy was triggered once (13c, 13d) for both Phi-4, DeepSeek, and gpt-4o:
% \textit{azure.core.exceptions.HttpResponseError: (content\_filter) The response was filtered due to the prompt triggering Microsoft's content management policy. Please modify your prompt and retry.}
% So, we tried to rerun the same questions. However, the error still appeared after two trials. 
% The corresponding question is "Er det etisk å genetisk modifisere mennesker?" asked in Norwegian (13c) and asked in Norwegian but requires the answer in English (question indexing starts from 0).
% Inner error: {
%     "code": "ResponsibleAIPolicyViolation",
%     "content_filter_result": {
%         "hate": {
%             "filtered": true,
%             "severity": "medium"
%         },

\subsection{Word count analysis}

\begin{figure}[!hbt]
    \centering
    \includegraphics[width=1\linewidth]{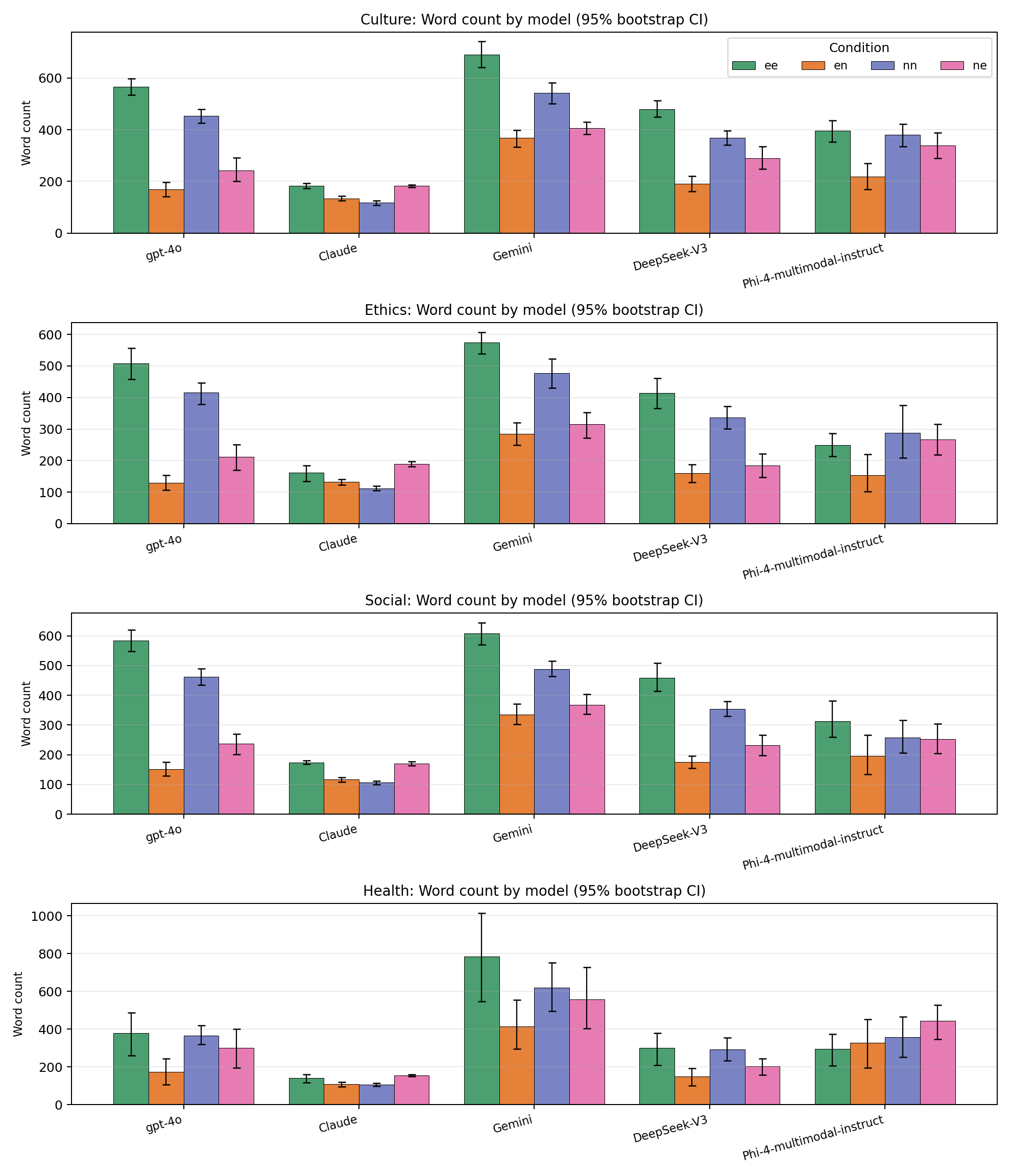}
    \caption{Mean word count (95\% bootstrap CI) by model and domain on the balanced dataset ($n=1{,}348$).}
    \label{fig:word_count_errorbars_by_model_by_topic}
\end{figure}

Across topics and models, response length follows a consistent ordering: the $ee$ condition is generally longest, $en$ is generally shortest, and $nn$/$ne$ fall between them. Figure~\ref{fig:word_count_errorbars_by_model_by_topic} shows this pattern by model and domain on the balanced 337-question dataset, with 95\% bootstrap confidence intervals.

\textit{Translation compression in three length units.} On the balanced dataset, the mean $ee \to en$ reduction is 52\% in words (427.8 $\to$ 203.8), 54\% in characters (3072.9 $\to$ 1419.6), and only 25\% in tokens (614.1 $\to$ 459.9) under the \texttt{cl100k\_base} tokenizer. Strikingly, $nn$ responses are 31\% \emph{longer} in tokens than $ee$ responses (804.3 vs.\ 614.1) despite being 20\% shorter in words. The word/character metrics track one another closely, but the token gap is much smaller, reflecting Norwegian's higher subword fertility on English-centric tokenizers. We therefore treat the word-based ``compression'' figure as an upper bound on how much shorter Norwegian responses truly are in the model's own units.

Overall, the $ee \to en$ length gap corresponds to a large effect size ($d\approx 1.32$ on the balanced dataset), but this pooled value hides considerable per-model heterogeneity (Table~\ref{tab-per-model-d}): Phi-4 sits just above the medium/large threshold ($d=0.78$), Claude is large ($d=1.52$), and GPT-4o is exceptionally large ($d=4.42$). The pooled $d$ is driven primarily by the more verbose models.

\begin{table}[!hbt]
\centering
\caption{Per-model Cohen's $d$ for the $ee$--$en$ length gap on the 68 non-translation questions. $n=68$ per condition per model.}
\label{tab-per-model-d}
\begin{tabular}{lcccc}
\toprule
Model & Mean $ee$ & Mean $en$ & $d$ & Interpretation \\
\midrule
GPT-4o                    & 542.5 & 157.1 & 4.42 & very large \\
DeepSeek-V3               & 440.9 & 175.3 & 3.11 & very large \\
Gemini~2.5~Pro            & 661.3 & 345.8 & 2.18 & very large \\
Claude~3.5~Haiku          & 170.9 & 126.2 & 1.52 & large \\
Phi-4-multimodal-instruct & 324.0 & 213.5 & 0.78 & medium-to-large \\
\bottomrule
\end{tabular}
\end{table}

\begin{figure}[!hbt]
    \centering
    \includegraphics[width=1\linewidth]{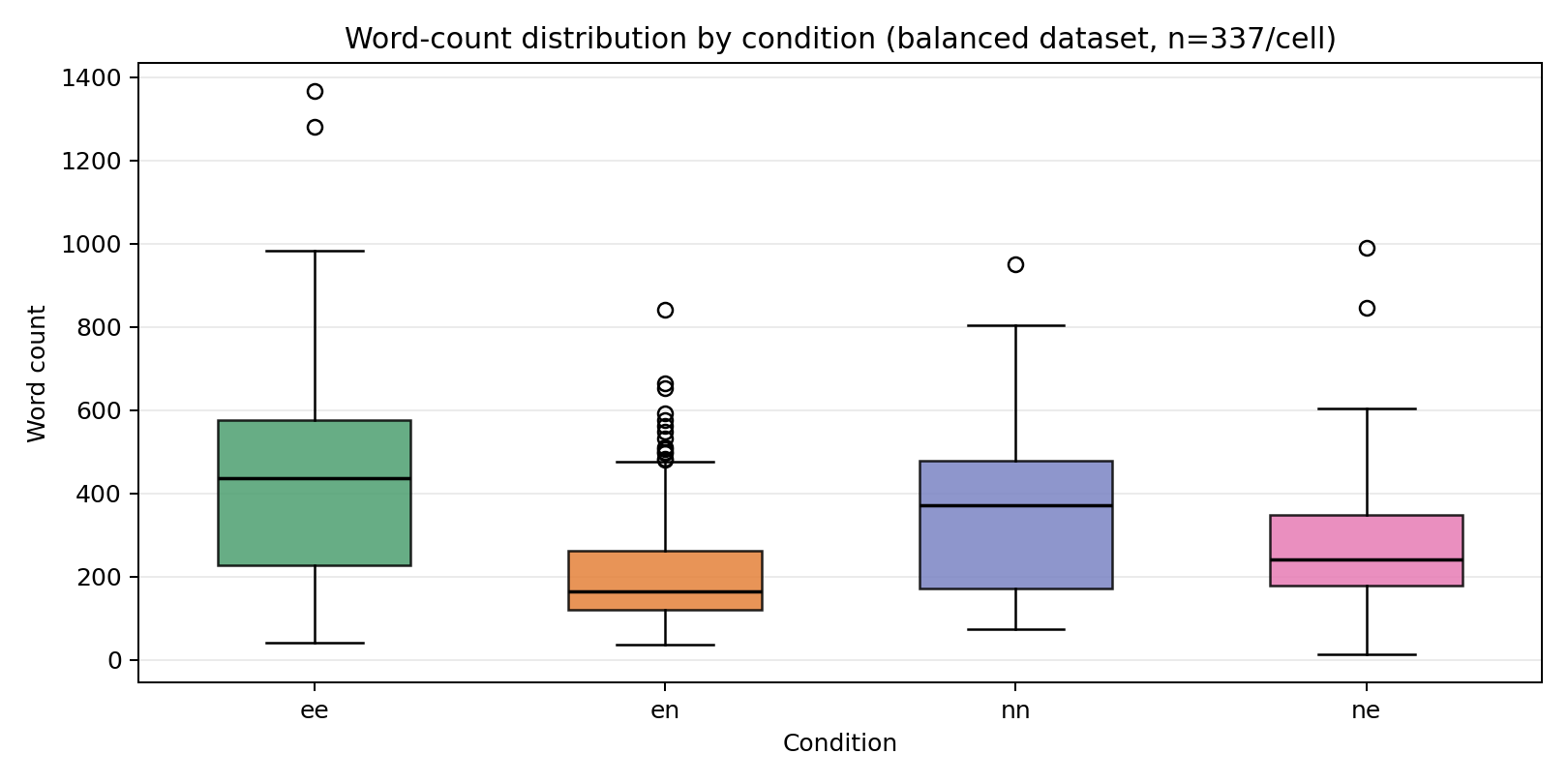}
    \caption{Word-count distributions by condition on the balanced dataset ($n=337$ per cell).}
    \label{fig:word_count_distribution_by_mode}
\end{figure}

Figure~\ref{fig:word_count_distribution_by_mode} shows the aggregate distributional pattern by condition.

%         mean     std  min   max  count
% Mode                                  
% ee    417.63  210.84   25  1366    350
% en    199.93  124.03   25   841    350
% ne    268.22  126.63   13   991    337
% nn    344.55  168.63   73   951    337

% Mean ± StdDev by Mode:
%   ee: 417.63 ± 210.84
%   en: 199.93 ± 124.03
%   ne: 268.22 ± 126.63
%   nn: 344.55 ± 168.63

% \begin{table}
% \caption{Word counts for the questions in the ethics topic}
% \label{tab-ethics-word-count}
% \begin{tabular}{lcccc}
% \toprule
%  & ee & en & nn & ne \\
% \midrule
% gpt & 508.20$\pm$117.39 & 129.70$\pm$54.68 & 415.00$\pm$76.98 & 211.68$\pm$88.79 \\
% deepseek & 414.40$\pm$113.75 & 159.30$\pm$68.04 & 337.11$\pm$78.76 & 183.84$\pm$84.04 \\
% gemini & 574.50$\pm$81.65 & 285.40$\pm$78.82 & 476.90$\pm$109.31 & 315.70$\pm$91.44 \\
% claude & 161.25$\pm$59.43 & 131.75$\pm$19.41 & 112.25$\pm$16.18 & 189.70$\pm$19.30 \\
% phi & 249.55$\pm$86.21 & 152.85$\pm$140.91 & 287.32$\pm$193.63 & 266.21$\pm$112.83 \\
% \bottomrule
% \end{tabular}
% \end{table}

Norwegian-output conditions ($en$, $nn$) are generally more concise (in words) than English-output conditions ($ee$, $ne$).

\subsection{Decomposing the Prompt and Response Language Effects}
Because the four conditions form a $2\times2$ design (prompt language $\times$ response language), we can separate the contribution of the \emph{prompt} language from that of the \emph{response} language rather than treating ``cross-lingual'' as a single factor. Table~\ref{tab-2x2} reports mean word counts for the four cells on the balanced dataset.

\begin{table}[!hbt]
\centering
\caption{Mean word count by prompt $\times$ response language on the balanced dataset ($n=337$ per cell across five models and four domains).}
\label{tab-2x2}
\begin{tabular}{lcc}
\toprule
 & Response: English & Response: Norwegian \\
\midrule
Prompt: English   & 427.8 ($ee$) & 203.8 ($en$) \\
Prompt: Norwegian & 268.2 ($ne$) & 344.6 ($nn$) \\
\bottomrule
\end{tabular}
\end{table}

Two patterns emerge. First, the effect of response language on length \emph{reverses} depending on the prompt language: under an English prompt an English response is on average about 224 words longer than a Norwegian one ($ee$ vs.\ $en$), whereas under a Norwegian prompt an English response is about 76 words \emph{shorter} than a Norwegian one ($ne$ vs.\ $nn$). Fitting the mixed-effects model $\text{words} \sim \text{prompt} \times \text{response} + \text{model} + (1\mid\text{question})$ on the analyzed set ($n=1{,}354$) yields a large and highly significant prompt$\times$response interaction (coefficient $+300.7$ words, $z=25.62$, $p<10^{-100}$), with a substantial question-level variance component (Group Var $\approx 2{,}483$) that a fixed-effects ANOVA would incorrectly pool into residual variance and thereby overstate the significance of. The main effects go in the expected directions (Norwegian prompt shortens: $-159.7$, $z=-19.24$; Norwegian response shortens: $-224.3$, $z=-27.09$). Refitting on $\log(1 + \text{words})$ as a robustness check for the right-skewed distribution yields the same qualitative and quantitative story (interaction $z=23.01$; back-transformed percentage reductions within 1--2 pp of the direct-means values in Table~\ref{tab-2x2}); see \ref{appendix-logscale}.

Second, the single most compressed condition is $en$, not cross-lingual generation in general. We therefore interpret ``translation compression'' as a prompt$\times$response interaction rather than a uniform property of cross-lingual prompting: the prompt language modulates how strongly the response language affects output length.

\subsection{Semantic Similarity and Cross-lingual Keyword Overlap}

\begin{figure}[!hbt]
    \centering
    \includegraphics[width=1\linewidth]{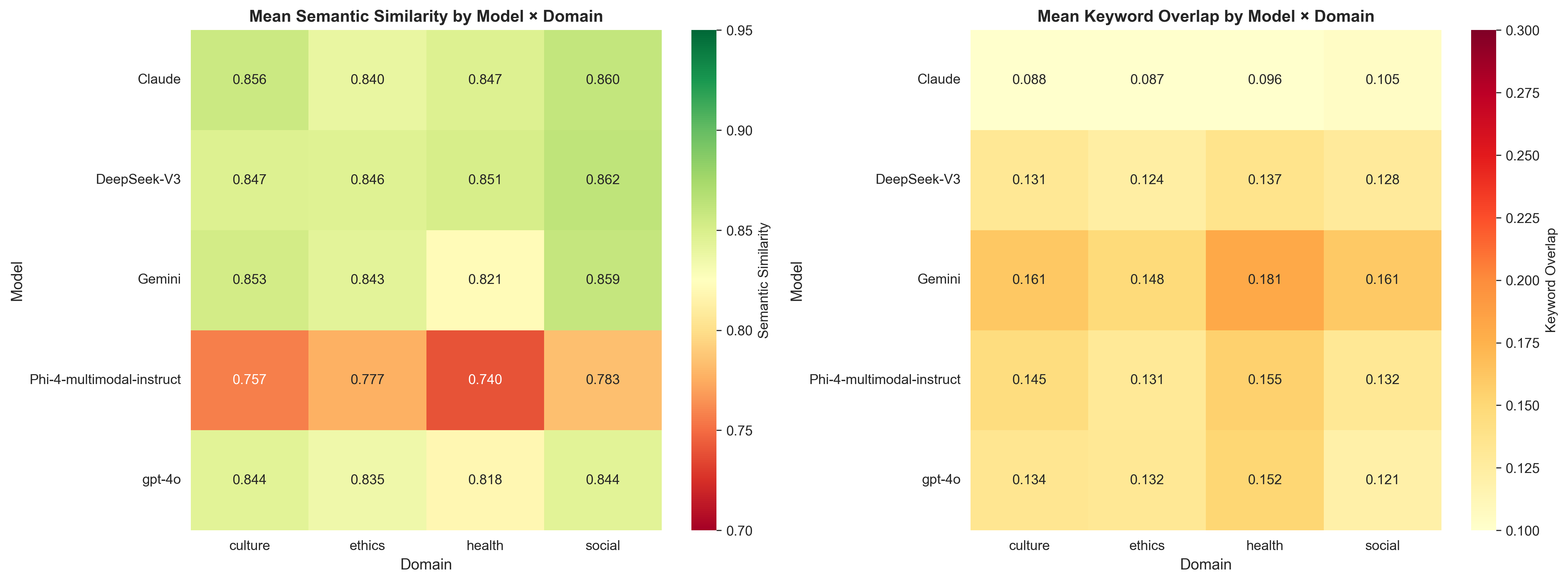}
    \caption{Mean semantic similarity (left) and mean raw-Jaccard keyword overlap (right) by model and domain, averaged over the six pairwise combinations of $\{ee, en, nn, ne\}$.}
    \label{fig:semantic_keyword_heatmaps}
\end{figure}

% Mean Semantic Similarity:
% domain                     culture  ethics  health  social
% model                                                     
% Claude                       0.856   0.840   0.847   0.860
% DeepSeek-V3                  0.847   0.846   0.851   0.862
% Gemini                       0.853   0.843   0.821   0.859
% Phi-4-multimodal-instruct    0.757   0.777   0.740   0.783
% gpt-4o                       0.844   0.835   0.818   0.844

% Mean Keyword Overlap (fixed TF-IDF):
% domain                     culture  ethics  health  social
% model
% Claude                       0.088   0.087   0.096   0.105
% DeepSeek-V3                  0.131   0.124   0.137   0.128
% Gemini                       0.161   0.148   0.181   0.161
% Phi-4-multimodal-instruct    0.145   0.131   0.155   0.132
% gpt-4o                       0.134   0.132   0.152   0.121

\begin{figure}[!hbt]
    \centering
    \includegraphics[width=1\linewidth]{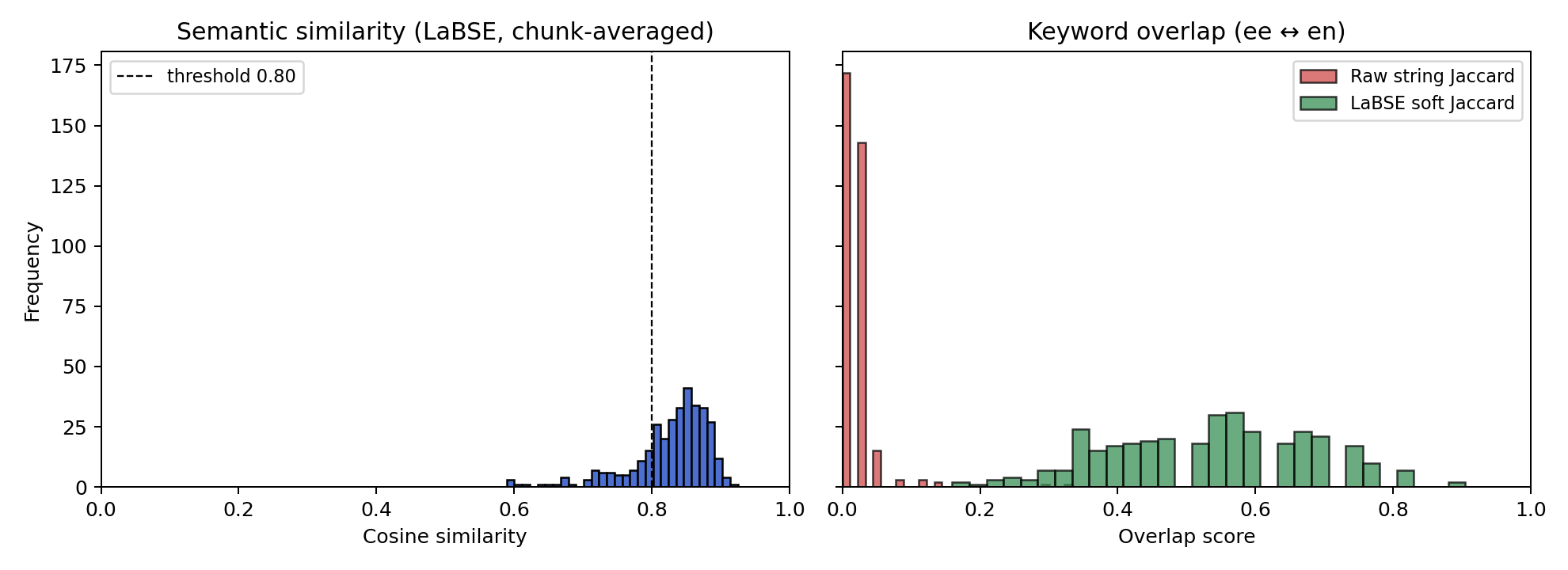}
    \caption{Left: distribution of chunk-averaged LaBSE cosine similarity across questions. Right: distribution of ee$\leftrightarrow$en keyword overlap under raw Jaccard (red) and LaBSE soft Jaccard (green). Raw Jaccard collapses near zero because English and Norwegian keyword surface forms rarely match; soft Jaccard reveals substantial concept-level overlap ($\approx 0.53$ mean).}
    \label{fig:semantic_lexical_paradox}
\end{figure}

% Semantic Similarity (Cosine): 0.832 ± 0.052
%   Range: [0.588, 0.916]
%   Above threshold (0.80): 274 / 337 (81.3%)

% Keyword Overlap (Jaccard): 0.129 ± 0.034
%   Range: [0.057, 0.225]
%   Above threshold (0.30): 0 / 337 (0.0%)

% PARADOX: High semantic + Low lexical overlap:
%   274 cases (81.3%)
%   Semantic-Lexical decoupling: CONFIRMED

\textit{Semantic stability with genuine paraphrasing.} Figure~\ref{fig:semantic_keyword_heatmaps} shows that mean LaBSE cosine similarity remains high across most model--domain combinations (roughly $0.82$--$0.86$, with lower values for Phi-4). Chunking long English responses at 250-word windows and re-embedding gives a near-identical mean (0.827 vs.\ 0.832 naive; per-question $\Delta = -0.005$ on the balanced dataset), so the LaBSE truncation concern is negligible in practice.

The lexical picture is only meaningful once we account for language. Restricting to the cross-lingual $ee \leftrightarrow en$ pair on the analyzed set ($n=340$ question-pairs), the mean raw Jaccard is $0.017$: because Norwegian and English keyword surface forms overlap almost never, this number essentially measures ``the two responses are in different languages'' and cannot diagnose paraphrasing. Under LaBSE soft Jaccard the same pairs score $0.530$ on average---a $+0.512$ lift. This raw--soft gap (Figure~\ref{fig:semantic_lexical_paradox}, right panel) provides the actual evidence for the paraphrasing interpretation: the models are reusing the same concepts, expressed in different surface vocabulary, and the reuse only shows up once keywords are mapped through a shared multilingual embedding.

\subsection{Information Density (Length--Vocabulary Companion)}

\begin{figure}[!hbt]
    \centering
    \includegraphics[width=\linewidth]{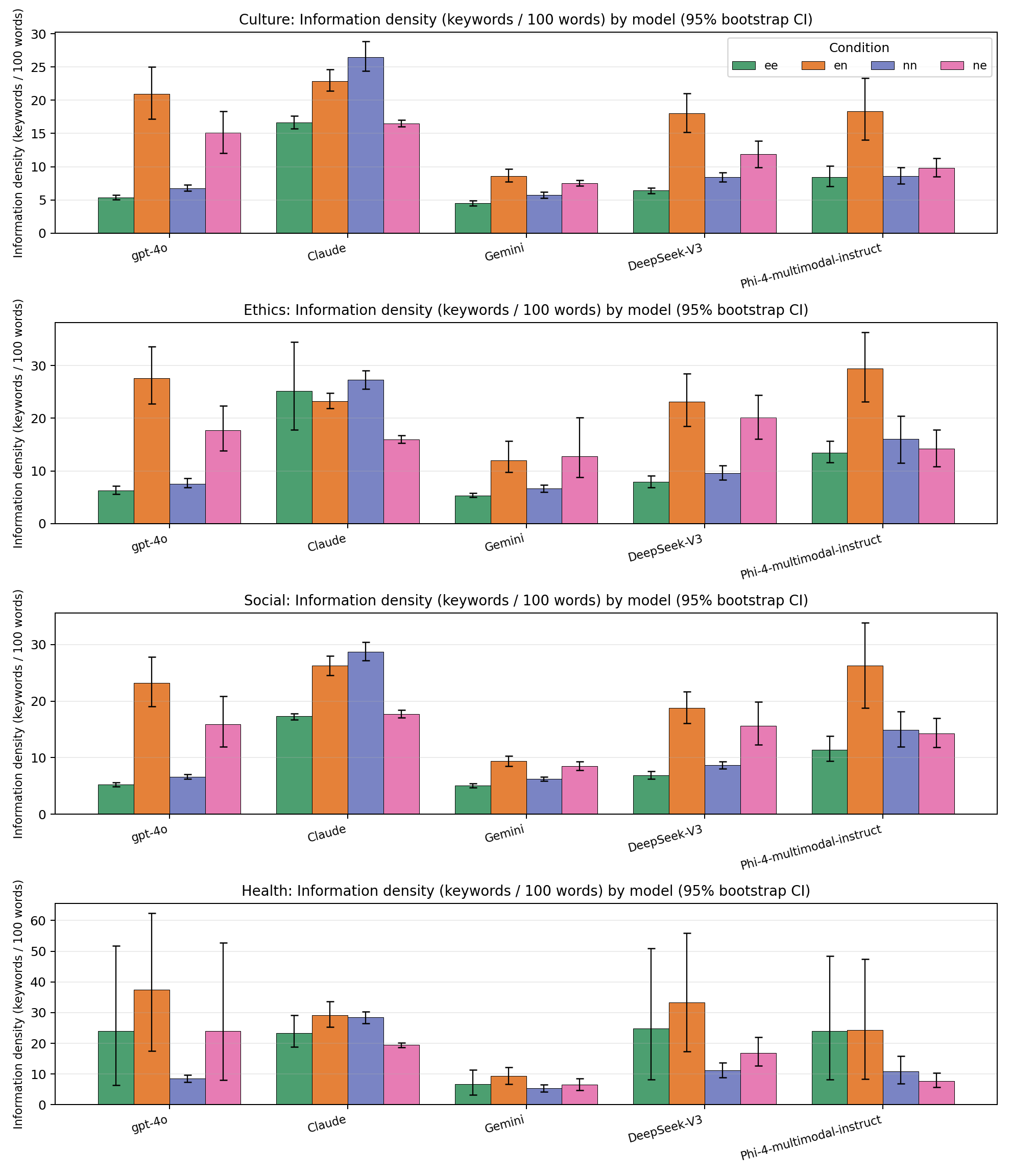}
    \caption{Mean information density (95\% bootstrap CI) by model and domain under each condition. CIs replace the earlier $\pm$SD bars so the intervals stay in the non-negative range.}
    \label{fig:information_density_by_model_by_domain}
\end{figure}

Information density is highest for $en$ across domains and most models (Figure~\ref{fig:information_density_by_model_by_domain}). The magnitude of the gap varies by model.

\begin{figure}[!hbt]
    \centering
    \includegraphics[width=1\linewidth]{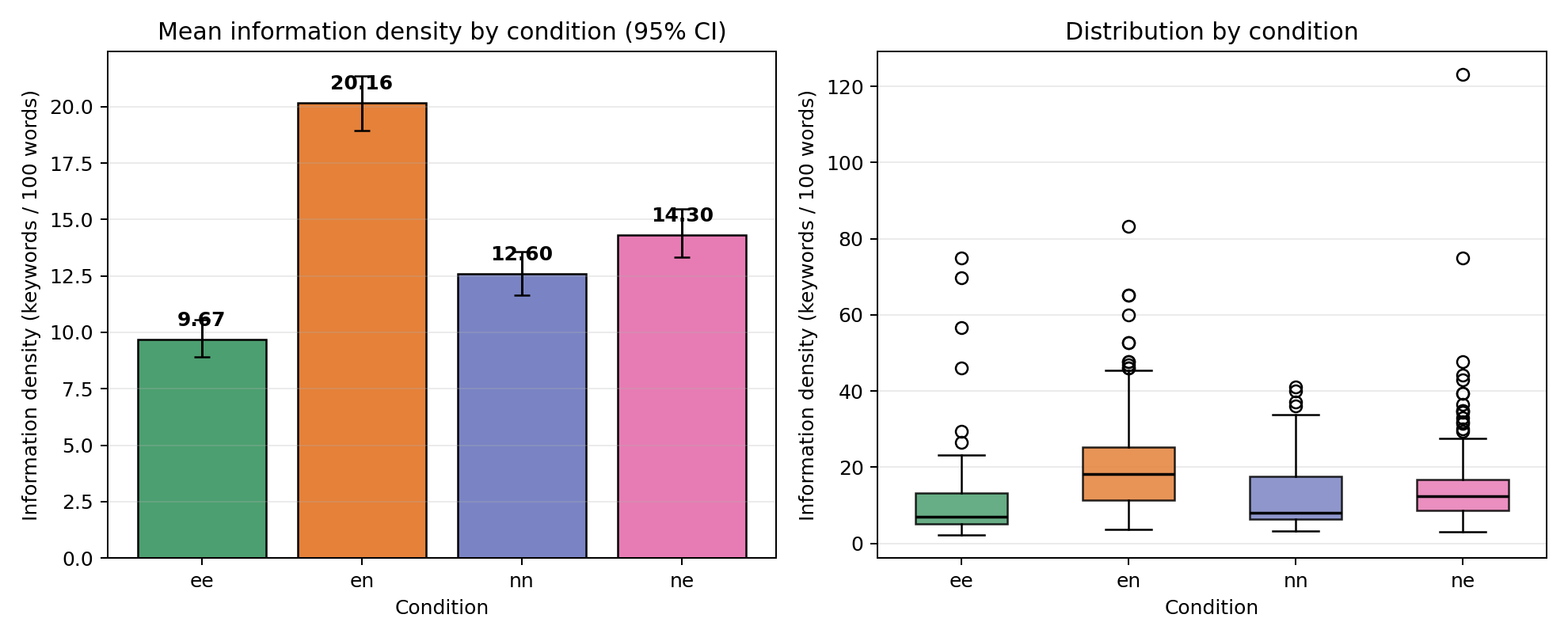}
    \caption{Information-density distributions by condition. Left: mean $\pm$ 95\% bootstrap CI. Right: full distribution.}
    \label{fig:information_density_dist}
\end{figure}

% Information Density by Variant:
% ============================================================
% Variant a (en):
%   Mean ± Std: 21.42 ± 14.90%
%   Count: 350

% Variant d (ne):
%   Mean ± Std: 14.30 ± 9.77%
%   Count: 337

% Variant c (nn):
%   Mean ± Std: 12.60 ± 8.93%
%   Count: 337

% Variant b (ee):
%   Mean ± Std: 11.25 ± 13.59%
%   Count: 350

As shown in Figure~\ref{fig:information_density_dist}, on the balanced dataset the mean information density by condition is highest for $en$ ($20.16$), followed by $ne$ ($14.30$), $nn$ ($12.60$), and $ee$ ($9.67$). Because the number of extracted keywords saturates near its cap for all but the shortest responses, this metric is a near-perfect monotonic inverse of length in our data (Spearman $\rho\approx -1.0$; Pearson $r\approx 0.98$ between density and $1/N_{\text{words}}$). It therefore re-expresses the length result from a complementary angle rather than constituting independent evidence of communicative ``efficiency,'' and we do not use it to argue for a separate quality gain.

Overall, the figure set indicates that language condition changes response length and lexical realization more strongly than semantic content. The key empirical pattern is high semantic alignment together with substantial cross-lingual paraphrasing (raw Jaccard $\approx 0.02$ vs.\ soft Jaccard $\approx 0.53$ for $ee \leftrightarrow en$), alongside a large length gap that shrinks by roughly half when measured in tokens.

\section{Discussion}\label{sec-discuss}
The core empirical picture from Section~\ref{sec-result} is:
\begin{itemize}
\item A large prompt$\times$response interaction on response length that reverses sign across cells: English prompts produce much longer English than Norwegian responses, while Norwegian prompts produce slightly shorter English than Norwegian ones.
\item A single-model story that is not uniform: per-model $d$ ranges from Phi-4 (0.78) to GPT-4o (4.42), so the pooled ``compression'' number is dominated by the more verbose models.
\item Semantic content is preserved across conditions ($\approx 0.83$ cosine similarity, unchanged under chunk-averaged embedding), while surface vocabulary is rearranged: cross-lingual raw Jaccard $\approx 0.02$ vs.\ LaBSE soft Jaccard $\approx 0.53$.
\end{itemize}
We interpret this as evidence that language choice reshapes \emph{how} content is packaged (length, surface vocabulary, lexical realization) more than \emph{what} is conveyed, but we are careful not to over-claim: none of our measurements directly test cultural or register content.

\subsection{Candidate Mechanisms}
We speculatively flag four mechanisms that plausibly drive the observed pattern; disentangling their relative contributions is beyond the scope of this study.

\paragraph{(1) Tokenizer fertility} English-centric BPE-family tokenizers segment Norwegian text into substantially more subword tokens per word than English text. Under the \texttt{cl100k\_base} reference tokenizer, mean $nn$ responses reach 804 tokens compared with 614 for $ee$ despite being shorter in words---so a portion of what looks like ``compression'' in word counts is a linguistic-typology and tokenizer effect rather than a decision made by the model. The truncation audit (\ref{appendix-truncation}) shows the 1{,}000-token cap was hit mostly on Norwegian outputs ($en$: 19, $nn$: 58) and rarely on long $ee$ responses ($ee$: 13), so the measured 52\% $ee$--$en$ word gap conflates model behavior with a truncation asymmetry between the two conditions. An uncapped rerun would be needed to separate the two.

\paragraph{(2) RLHF verbosity bias} Preference data for instruction-tuning is overwhelmingly English, and human raters have well-documented biases toward longer, more elaborative English answers~\cite{singhal2024longlonger}. The models we tested that are known to receive heavy English RLHF (GPT-4o, DeepSeek-V3, Gemini) show the largest $ee$--$en$ gaps ($d \ge 2.18$); Phi-4, with a smaller RLHF footprint, shows the smallest ($d=0.78$). This is consistent with, but does not prove, an RLHF-verbosity account.

\paragraph{(3) Pretraining-corpus asymmetry} Norwegian is a low-resource language relative to English in every publicly documented web-scale corpus. Under scarcity, models may plausibly default to more terse, less rhetorically elaborated Norwegian generations simply because Norwegian long-form prose is under-represented in their training distribution. This is compatible with the observation that Norwegian outputs also show higher density (fewer discourse markers per content word) even when produced by an English prompt.

\paragraph{(4) Safety hedging and refusals} The three items refused in this dataset were all refused only in Norwegian-prompt conditions, and health~Q9/Q10---explicit translation asks---were refused universally. This suggests safety-classifier behavior is itself language-conditioned, in a way that would deserve its own study. For length analyses we address this by dropping any question refused in any condition, but the pattern itself is a limitation of comparing generation behavior across languages naively.

\subsection{Limitations}
\paragraph{Token-cap truncation} Generation was capped based on the defaults (1{,}000 tokens (2{,}000 for Claude)), and stop reasons were not logged. A post-hoc audit finds 110 responses that end without terminal punctuation, concentrated in Norwegian-output conditions ($en$: 19, $nn$: 58; together 69\% of the truncated set) rather than in English responses ($ee$: 13, $ne$: 21). This is consistent with Norwegian tokenizer fertility exhausting the 1{,}000-token budget before the response completes---the same mechanism we invoke to explain the tokens-vs.-words gap. A rerun with an unlimited or much higher cap would be needed to fully rule out truncation contamination of the length effects, particularly of the $nn$ mean.

\paragraph{Density is a length proxy} The information-density metric is bounded by a fixed keyword cap and is consequently a near-perfect monotonic inverse of response length (Spearman $\rho \approx -1$). It should be read as a descriptive companion to the word-count analysis, not as an independent efficiency measure.

\paragraph{Cultural framing not quantified} Register and framing differences (``more elaborative,'' ``more direct'') would need dedicated measures---e.g., formality classifiers, hedging-phrase frequency, or Norway-specific named-entity counts---to be established quantitatively. We do not make quantitative cultural claims in this paper.

\paragraph{Scope} The study covers a single language pair (English--Norwegian, Bokmål), five models, and 68 non-translation questions per model. Generalization to other language pairs, Nynorsk, larger question sets, and multi-turn settings remains to be verified. Question provenance is also mixed (Tulu, Copilot, and manual authoring across topics); we have not tested whether provenance systematically affects any of the reported metrics.

\section{Conclusion}\label{sec-conclu}
Prompt and response language are not neutral parameters: they jointly shape output length, lexical realization, and---to a degree we do not attempt to quantify here---rhetorical framing. Concretely, three findings survive the robustness checks we ran. First, the $ee \to en$ length reduction is $\sim 52\%$ in words but only $\sim 25\%$ in tokens, so a large portion of the ``compression'' effect reflects Norwegian tokenizer fertility rather than the model producing genuinely more compact content; the interaction between prompt and response language is nonetheless large and highly significant in a mixed-effects model that respects the nested structure of the design ($z = 25.6$). Second, semantic content is preserved across conditions ($\approx 0.83$ cosine similarity, unchanged under chunk-averaged embedding), while cross-lingual keyword overlap moves from $\approx 0.02$ under raw Jaccard to $\approx 0.53$ under LaBSE soft Jaccard---direct evidence of concept-level paraphrasing that is invisible to surface-string comparison. Third, the pooled effect size is heterogeneous across models (per-model $d$ ranges from 0.78 to 4.42), so a single ``compression'' number should be reported alongside its per-model breakdown. For multilingual prompt engineering, the practical takeaway is that changing the response language of a query changes not only its language but its length budget, tokenization cost, and lexical realization; downstream systems that rely on any of these should test in the target-language setting rather than extrapolating from English behavior.

\appendix
\section{Lists of Experimental Questions}
\label{appendix-questions}

\subsection{Ethics questions}
\begin{enumerate}
    \item What is ethics?
    \item What is the difference between ethics and morality?
    \item Is it ever ethical to lie?
    \item How do ethical theories guide decision-making?
    \item What are the principles of utilitarianism?
    \item What are the challenges of ethical relativism?
    \item What role does culture play in ethics?
    \item Should animals have the same rights as humans?
    \item How can businesses act ethically?
    \item Can ethics exist without religion?
    \item What are some ethical issues in technology?
    \item How can ethics address environmental challenges?
    \item What is the ethical dilemma of self-driving cars?
    \item Is it ethical to genetically modify humans?
    \item What makes an action good or bad?
    \item What are the main criticisms of deontological ethics?
    \item How do we define justice in ethical terms?
    \item What is the importance of ethical leadership?
    \item Can ethical behavior be taught?
    \item How do personal values influence ethical decisions?
\end{enumerate}
\subsection{Cultural questions}
\begin{enumerate}
    \item What are some significant cultural festivals celebrated around the world?
    \item Describe the influence of the Renaissance on European culture.
    \item How does traditional clothing vary across different cultures?
    \item What role does storytelling play in preserving indigenous cultures?
    \item Explain the concept of cultural appropriation and its implications.
    \item How do food traditions reflect cultural identity?
    \item Describe the significance of the tea ceremony in Japanese culture.
    \item What are the key features of the Maori culture in New Zealand?
    \item How does architecture in different countries reflect cultural values?
    \item What are some traditional musical instruments unique to specific cultures?
    \item How do wedding traditions differ across cultures?
    \item What is the significance of the Holi festival in India?
    \item Describe the role of masks in African traditional ceremonies.
    \item What is the meaning and importance of the Dreamtime in Aboriginal Australian culture?
    \item How does the Carnival of Brazil reflect the country's cultural diversity?
    \item What are some traditional dances from the Middle East?
    \item Describe the influence of Confucianism on Chinese culture.
    \item How does the culture of the Inuit people adapt to Arctic life?
    \item What are some traditional storytelling forms in Native American cultures?
    \item Describe the significance of the cherry blossom season in Japan.
\end{enumerate}
\subsection{Health-related questions (analyzed set, $n=8$)}
\begin{enumerate}
    \item What are the guidelines for using infant formula when breastfeeding isn't possible? What brands are recommended?
    \item What is the standard treatment for an ankle fracture? How long is the recovery process?
    \item I'm having persistent headaches. How do I get a referral to a neurologist?
    \item What should I do if I think someone is having a stroke? What emergency services should I contact?
    \item What are the most effective contraception methods for someone in their 20s? How do I access them?
    \item Should I take vitamin D supplements during winter? What's the recommended dosage?
    \item How do I properly measure my blood pressure at home? What equipment do I need?
    \item What resources are available for someone experiencing depression? Where can I find professional help?
\end{enumerate}
\subsection{Social related questions}
\begin{enumerate}
    \item What is the role of social media in shaping public opinion?
    \item How does peer pressure impact decision-making?
    \item What are the benefits and drawbacks of social networks?
    \item How does cultural diversity affect social interactions?
    \item What is the importance of empathy in social relationships?
    \item How do stereotypes influence societal behavior?
    \item What are the challenges of building inclusive communities?
    \item How can social inequality be addressed effectively?
    \item What is the impact of technology on social connections?
    \item How does urbanization influence social dynamics?
    \item What role does education play in promoting social cohesion?
    \item What are the ethical considerations of social experiments?
    \item How do social movements bring about change?
    \item What are the effects of globalization on social structures?
    \item How do social norms evolve over time?
    \item What is the impact of generational differences on social values?
    \item How can individuals contribute to societal well-being?
    \item What are the consequences of social isolation?
    \item How does social trust influence societal development?
    \item What are the roles of family in shaping social values?
\end{enumerate}

\section{Prompt Templates}\label{appendix-prompts}
Each question was sent as a single user turn under one of the four conditions below. The Azure-hosted models (GPT-4o, DeepSeek-V3, Phi-4-multimodal-instruct) and Claude were sent the system message ``You are a helpful assistant.''; Gemini was called via \texttt{generate\_content} with no explicit system prompt (SDK default). Placeholders \texttt{\{Q\}$_{\text{EN}}$} and \texttt{\{Q\}$_{\text{NO}}$} hold the English or Norwegian version of the question.
\begin{itemize}
    \item \textbf{$ee$ (English $\to$ English):} \texttt{\{Q\}$_{\text{EN}}$}
    \item \textbf{$en$ (English $\to$ Norwegian):} \texttt{answer in Norwegian:\{Q\}$_{\text{EN}}$}
    \item \textbf{$nn$ (Norwegian $\to$ Norwegian):} \texttt{\{Q\}$_{\text{NO}}$}
    \item \textbf{$ne$ (Norwegian $\to$ English):} \texttt{svar p{\aa} engelsk:\{Q\}$_{\text{NO}}$}
\end{itemize}
Two design choices are worth flagging explicitly, since they mean the four conditions are not a clean $2\times2$ of language alone. First, $ee$ and $nn$ carry no explicit language instruction (relying on the implicit convention that a model answers in the language of the query), whereas $en$ and $ne$ prepend an explicit cue---so any $ee$--$en$ or $nn$--$ne$ difference partially reflects the presence of the instruction. Second, the two cues are themselves in different languages: $en$'s cue is in English (``answer in Norwegian:''), while $ne$'s cue is in Norwegian (``svar p{\aa} engelsk:''). Effects attributed to prompt language in the cross-lingual conditions therefore combine (a) the primary question's language, (b) the presence of a cue, and (c) the cue's own language.

\section{Translation-task health items (excluded from length analyses)}\label{appendix-translation}
These two items were part of the health topic but ask explicitly for translation, which makes the response language over-determined and (in Norwegian-prompt conditions) triggered universal refusals. They are excluded from the length, density, similarity, and Jaccard analyses reported in the main text and are shown here for completeness.
\begin{enumerate}
    \setcounter{enumi}{8}
    \item Please translate this medical report excerpt into Norwegian, maintaining the technical accuracy: ``The cardiovascular assessment revealed moderate aortic regurgitation with an ejection fraction of 56\%. Pharmacological management includes ACE inhibitors alongside diuretic therapy. Regular monitoring of renal function is indicated.''
    \item Please translate this patient's explanation into Norwegian, keeping it easy to understand: ``Your heart test showed that one of your heart valves lets some blood flow backward. Your heart is still pumping well. You'll need to take two types of medicine: one to help your blood vessels relax, and another to reduce fluid buildup. We'll need to check your kidney function regularly.''
\end{enumerate}

\section{Refusal and Truncation Audits}\label{appendix-refusals}
\label{appendix-truncation}
On the raw 1{,}400 intended responses (5 models $\times$ 70 questions $\times$ 4 conditions), 26 were declined by content filters, all in Norwegian-prompt conditions. Table~\ref{tab-refusals} summarizes; balancing the design by dropping any question refused in \emph{any} condition removes 13 (model, question) pairs and leaves 1{,}348 responses (337 per condition).
\begin{table}[!hbt]
\centering
\caption{Number of missing responses per model $\times$ condition.}
\label{tab-refusals}
\begin{tabular}{lcccc}
\toprule
Model & $ee$ & $en$ & $nn$ & $ne$ \\
\midrule
Claude~3.5~Haiku          & 0 & 0 & 2 & 2 \\
DeepSeek-V3               & 0 & 0 & 3 & 3 \\
Gemini~2.5~Pro            & 0 & 0 & 2 & 2 \\
Phi-4-multimodal-instruct & 0 & 0 & 3 & 3 \\
GPT-4o                    & 0 & 0 & 3 & 3 \\
\bottomrule
\end{tabular}
\end{table}

A separate post-hoc audit of stop reasons (which the collection scripts did not log) finds 110 raw response files that terminate without terminal punctuation. Table~\ref{tab-truncation} shows the breakdown by model and condition: truncations concentrate in Norwegian-output conditions ($en$: 19, $nn$: 58; together 69\% of the total), not in English ones ($ee$: 13, $ne$: 21). Phi-4 (47 truncations) and Claude (21) account for the majority, and DeepSeek-V3 shows zero truncations in every condition except $nn$. This pattern is consistent with Norwegian's higher subword-tokenizer fertility exhausting the fixed 1{,}000-token \texttt{max\_tokens} cap before generation completes. We flag this as the most important open question in the current dataset: a rerun with an unlimited or much higher cap and logged stop reasons would be required to fully rule out truncation contamination---particularly of the $nn$ mean length.

\begin{table}[!hbt]
\centering
\caption{Number of responses terminating without terminal punctuation, by model and condition.}
\label{tab-truncation}
\begin{tabular}{lccccc}
\toprule
Model & $ee$ & $en$ & $nn$ & $ne$ & Total \\
\midrule
Claude~3.5~Haiku          & 2 & 7  & 11 & 1  & 21 \\
DeepSeek-V3               & 0 & 0  & 13 & 0  & 13 \\
Gemini~2.5~Pro            & 3 & 1  & 1  & 5  & 10 \\
Phi-4-multimodal-instruct & 2 & 10 & 21 & 14 & 47 \\
GPT-4o                    & 7 & 0  & 12 & 0  & 19 \\
\midrule
Total                     & 14 & 18 & 58 & 20 & 110 \\
\bottomrule
\end{tabular}
\end{table}

\section{Paired Example: Claude, health Q0 (infant formula)}\label{appendix-paired}
Both excerpts are the opening portion of the model's response to the same question in the same session. The $ee$ response recommends the US-market brands Similac, Enfamil, Gerber Good Start, and Earth's Best Organic; the $en$ response recommends the Norway-market brands Tine Nutrilett, Semper, Hipp, Aptamil, Nan, and Nutricia. Both list broadly the same clinical guidance (formula choice, preparation, feeding schedule), but with locally appropriate product and regulator references.

\paragraph{$ee$ (English $\to$ English, first $\sim$160 words).}
\begin{quote}\small
Here are some general guidelines for using infant formula:\\
1. Choosing a Formula: Use iron-fortified formula for infants under 12 months. Popular recommended brands include Similac, Enfamil, Gerber Good Start, Earth's Best Organic.\\
2. Preparation Guidelines: Always wash hands before preparing. Use sterile bottles and clean water. Follow mixing instructions precisely. Do not dilute formula to make it last longer. Discard unused formula after 1 hour at room temperature.\\
3. Feeding Recommendations: Feed on demand, typically every 2--3 hours. Use appropriate nipple size for baby's age. Hold baby semi-upright during feeding. Burp periodically during feeding.\\
4. Important Considerations: Consult pediatrician about specific formula needs\ldots
\end{quote}

\paragraph{$en$ (English $\to$ Norwegian, first $\sim$130 words, our English gloss in brackets).}
\begin{quote}\small
Her er retningslinjene for bruk av morsmelkerstatning n\aa r amming ikke er mulig [\textit{Here are the guidelines for using infant formula when breastfeeding is not possible}]:\\
Generelle retningslinjer: Konsult\'er alltid lege eller helsesykepleier f{\o}r valg av morsmelkerstatning. F{\o}lg anvisningene p\aa\ pakken n{\o}ye for korrekt blanding. Bruk rent vann og sterile flasker\ldots\\
Anbefalte merker i Norge [\textit{Recommended brands in Norway}]: Tine Nutrilett, Semper, Hipp, Aptamil, Nan, Nutricia.\\
Viktige tips [\textit{Important tips}]: Velg erstatning tilpasset barnets alder. Sjekk at produktet oppfyller norske helsemyndigheters krav [\textit{that the product meets Norwegian health-authority requirements}]\ldots
\end{quote}

\section{Log-scale robustness check for the length model}\label{appendix-logscale}
Because word counts are right-skewed (median $\approx 240$, maximum $\approx 1{,}370$), we refit the mixed-effects length model on $\log(1 + \text{words})$ as a robustness check. This makes the residual distribution more nearly symmetric and, more usefully, makes each fixed-effect coefficient interpretable as a \emph{multiplicative} change on the raw scale via $\exp(\hat{\beta})$. Table~\ref{tab-logscale} reports the fixed effects on both scales; results reproduce the raw-scale narrative to within a percentage point and are qualitatively identical.

\begin{table}[!hbt]
\centering
\caption{Mixed-effects length model on $\log(1 + \text{words})$, analyzed set ($n=1{,}354$; $\text{words} \sim \text{prompt} \times \text{response} + \text{model} + (1\mid \text{question})$). Percentages are back-transformed as $1 - \exp(\hat{\beta})$ and represent the marginal effect at the reference level.}
\label{tab-logscale}
\begin{tabular}{lccccc}
\toprule
Term & $\hat{\beta}_{\log}$ & SE & $z$ & Back-transformed & Direct-means \\
\midrule
Intercept                       &  5.320 & 0.036 & 146.20 & --- & --- \\
Prompt: Norwegian               & $-0.443$ & 0.030 & $-14.91$ & $-36\%$ (ee $\to$ ne) & $-37\%$ \\
Response: Norwegian             & $-0.764$ & 0.030 & $-25.79$ & $-53\%$ (ee $\to$ en) & $-52\%$ \\
Prompt$\times$Response (NO,NO)  & $+0.966$ & 0.042 & $+23.01$ & --- & --- \\
\bottomrule
\end{tabular}
\end{table}

The interaction is large and highly significant on the log scale ($z=23.01$, comparable to the raw-scale $z=25.62$). Combining the main effects and the interaction back-transforms to $\exp(-0.443 - 0.764 + 0.966) - 1 \approx -21\%$ for the nn cell relative to ee (direct-means value: $-19\%$). Direction, significance, and magnitude of every effect match the raw-scale fit; we retain the raw-scale numbers in the main text because Table~\ref{tab-2x2} and Table~\ref{tab-per-model-d} are already expressed in word counts.

\section{Data and Code Availability}\label{appendix-availability}
All response files, analysis code, and figure-generation scripts are on \hyperlink{https://github.com/nhan123lise/prompt-language-effects-llm}{Github}. The reproduction pipeline (\texttt{functions/recompute\_b.py}) regenerates every number and figure in Sections~\ref{sec-eval-strategy}--\ref{sec-result} from the raw text responses. Queries were issued between March and June 2026. Exact model identifiers used: \texttt{gpt-4o} (Azure snapshot \texttt{2024-08-06}), \texttt{claude-3-5-haiku-latest}, \texttt{gemini-2.5-pro-preview-03-25}, \texttt{DeepSeek-V3} (Azure OpenAI-compatible endpoint), \texttt{Phi-4-multimodal-instruct} (Azure OpenAI-compatible endpoint). Decoding parameters were held at API defaults except \texttt{max\_tokens}: 1{,}000 for the Azure-hosted models, 2{,}000 for Claude, and unspecified for Gemini. No sampling was performed (single call per question per condition). LaBSE embeddings use \texttt{sentence-transformers/LaBSE} at its default hyperparameters.

 \bibliographystyle{elsarticle-num}
 \bibliography{ref}

\end{document}